\documentclass[conference]{IEEEtran}  

\IEEEoverridecommandlockouts                              

\usepackage{amsmath} 
\usepackage{amssymb}  
\usepackage{booktabs}
\usepackage{multirow}
\usepackage{graphicx}
\usepackage{subfigure}
\usepackage{cite}
\usepackage{float}

\usepackage{tikz}
\usetikzlibrary{positioning} 
\usetikzlibrary{matrix}
\usetikzlibrary{shapes}
\usepackage{pgfplots}
\pgfplotsset{compat=1.15}
\usepackage{pgfplotstable}

\usepackage{multirow, makecell}
\usepackage{array}

\usepackage[normalem]{ulem}
\useunder{\uline}{\ul}{}
\usepackage{siunitx}
\usepackage{tabularx}
\usepackage{tabu}
\PassOptionsToPackage{hyphens}{url}
\usepackage{hyperref}
\usepackage{censor}

\title{
Context-free Self-Conditioned GAN for \\ Trajectory Forecasting
\thanks{This work was supported by the Wallenberg AI, Autonomous Systems and Software Program (WASP) funded by the Knut and Alice Wallenberg Foundation.}
\thanks{ \textbf{This is the accepted manuscript. The final published article is available at \url{https://doi.org/10.1109/ICMLA55696.2022.00196}}}%
\thanks{ \textbf{© 2022 IEEE.  Personal use of this material is permitted.  Permission from IEEE must be obtained for all other uses, in any current or future media, including reprinting/republishing this material for advertising or promotional purposes, creating new collective works, for resale or redistribution to servers or lists, or reuse of any copyrighted component of this work in other works.}}

}

\begin{document}

\author{\IEEEauthorblockN{Tiago Rodrigues de Almeida}
\IEEEauthorblockA{AASS, \"{O}rebro University, Sweden\\
tiago.almeida@oru.se}
\and
\IEEEauthorblockN{Eduardo Gutierrez Maestro}
\IEEEauthorblockA{AASS, \"{O}rebro University, Sweden\\
eduardo.gutierrez-maestro@oru.se}
\and
\IEEEauthorblockN{Oscar Martinez Mozos}
\IEEEauthorblockA{AASS, \"{O}rebro University, Sweden\\
oscar.mozos@oru.se}
}
\maketitle

\begin{abstract}
In this paper, we present a context-free unsupervised approach based on a self-conditioned GAN to 
learn different modes from 2D trajectories. Our intuition is that each mode indicates a different 
behavioral moving pattern in the discriminator's feature space. We apply this approach to the 
problem of trajectory forecasting. We present three different training settings based on 
self-conditioned GAN, which produce better forecasters. We test our method in two data sets: 
human motion and road agents. Experimental results show that our approach outperforms 
previous context-free methods in the least representative supervised labels while performing 
well in the remaining labels. In addition, our approach outperforms globally in human motion, 
while performing well in road agents.
\end{abstract}
\section{INTRODUCTION}

Trajectories forectasing in the 2D space is a problem that has increasingly gained more focus inside the research community~\cite{ridel18, rudenkosurv20}. Example applications include surveillance systems~\cite{zhou15} and autonomous driving that require reliable technologies for the motion forecasting task~\cite{lorenzo20,cai21}. In addition, service robots need safer and more accurate people's trajectory forecasters to provide better human-robot experiences~\cite{roberts20, antonucci22}.

In this paper, we address the problem of trajectory forecasting using the information from the observed initial steps. Contrary to other approaches that use extra context information around the moving agent~\cite{alahi16,gupta18, sadeghian19,kosaraju19,jiachen19,dendorfer20,kothari21}, our method is context-free and only uses the trajectory information. The reason for that is two-fold. First, we want our approach to be flexible and useful in different environments. Second, and inspired by~\cite{scholler20}, it is yet not clear the use of social and context features for the trajectory prediction task. 

Predicting trajectories based solely on the observed track is a most challenging problem since the agent can have a panoply of different behaviors. However, similar situations can lead to akin behavioral patterns. Additionally, from the human motion perspective, people's movement encompasses different forms~\cite{templeton18}. Thus, we go beyond existing works by asking: is it possible to model similar motion patterns? Can these help the motion predictor 
to model a more diversified behavioral distribution? The latter question is explicitly connected to the fact that current approaches are biased toward the dominant 
behavior in the input distribution. Therefore, the hypothesis is that capturing different modes in the data may provide meaningful signals to learn a broader 
range of motion patterns.

Currently, state-of-the-art approaches are mainly centered on generative modeling~\cite{gupta18, sadeghian19, kosaraju19, jiachen19, dendorfer20, kothari21}. That is due to the 
capability of generating multiple tracks for one single observed trajectory. However, Generative Adversarial Networks (GANs) undergo the so-called mode collapse problem~\cite{akash17}, in which the generator may not be able to model the least dominant modes when the input distribution is biased to a specific mode. To overcome this problem, the work in~\cite{liu20} leveraged a framework based on a self-conditioned GAN to generate images conditioned on an unsupervised label. The idea is to capture different modes represented by clusters formed in the discriminator's feature space, thus facilitating the generation of more diverse images. 

In this paper, we adapt the self-conditioned GAN~\cite{liu20} to the 2D trajectory domain and apply it to the problem of trajectory forecasting. Moreover, we present three training settings based on self-conditioned GAN, which produce better forecasters. We test our method in two data sets: the TH\"{O}R~\cite{rudenko20} data set, which contains human motion trajectories; and the Argoverse~\cite{ming19} data set, which contains different road agents trajectories. Experimental results show that our approach outperforms previous context-free methods in the least representative supervised labels of these data sets while performing well in the remaining labels. In addition, our approach outperforms globally in human motion, while performing well in road agents. Also, we present insights into the clusters resulting from the self-conditioned GAN. Finally, we provide a tool for preprocessing the TH\"{O}R data set, which is made available~\cite{pythor}.  

\begin{figure}[t]
    \centering
    \tikzset{myLine1/.style = {->, thick, >=stealth},
    mynarrownodes/.style = {node distance=0.75cm and 0.3cm},
    ball/.style={
        ellipse,
        minimum width=3cm,
        minimum height=1.5cm,
        draw
    },
    }
    
    \begin{tikzpicture}
        \tikzstyle{bigbox} = [draw=black!50, thick, fill=gray!20, rounded corners, rectangle]

        \node[align=center] (input_data) at (0,0) {Trajectories \\ data set};
        \node[bigbox, align=center, right=0.8cm of input_data](smgan) {Self-Conditioned \\ GAN};

        \node[circle,draw,fill=white, right=0.75cm of smgan, minimum size=1.5cm] (modes) {};
        \node[circle,draw,fill=lightgray, minimum size=0.2cm] at (6.2,0.4) (m1) {};
        \node[ellipse,minimum width=0.5cm,minimum height=0.15cm,draw,fill=lightgray] at (5.7,0.4) (m2) {};
        \node[ellipse,minimum width=0.4cm,minimum height=0.7cm,draw,fill=lightgray] at (6.2,-0.25) (m3) {};
        \node[ellipse, rotate=135, minimum width=0.7cm,minimum height=0.1cm,draw,fill=lightgray] at (5.6,-0.25) (m4) {};

        \node [align=center]at (4.4,-0.8) (assump){\emph{soft-assumptions}};
        \node [align=center, above=0.01cm of modes](modes_str){Modes};
        \node [align=center] (dummy) at(5.35, 0) {};
        \node [align=center, below =0.01cm of modes] (dummy_plus) {};
        \node[bigbox, align=center, below=0.8cm of smgan](gan) {Context Free \\ GAN};
        
        \node [align=center, right=1.0cm of gan](output){Forecasted \\ trajectory};
        \node [align=center, left=1.0cm of gan](initial_traj){Initial \\ track};

        \draw[myLine1] (input_data.east) to[out=0, in=180] (smgan.west);
        \draw[myLine1] (smgan.east) to[out=0, in=180] (dummy.west);
        \draw[myLine1] (gan.east) to[out=180, in=180] (output.west);

        \draw[myLine1] (modes.south) -- ++(0,-0.2) -|  (gan.north);
        \draw[myLine1] (initial_traj.east) to[out=0, in=180] (gan.west);	

    \end{tikzpicture}
    \caption{Proposed framework's overview. First, self-conditioned GAN learns the different modes of the input data. Then, this information is used as training settings (via \emph{soft-assumptions}) to improve the prediction in specific modes.
    }
    \label{Fig1:framework}
    \vspace{-4.5pt}
\end{figure}
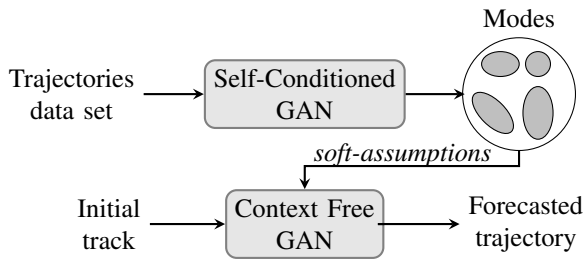
\section{RELATED WORK}

        




The motion prediction and analysis tasks can be divided into two main branches: context-free and context-aware approaches. The former focus its study uniquely on the observed motion steps. That is, in the scope of motion forecasting, context-free methods estimate the future locations based on the initial 
trace~\cite{becker18}. Besides the initial part of the trajectory, the latter approaches also consider the context. 
Mainly, the context is social interactions between agents~\cite{alahi16, gupta18, kothari21} and the scene's visual context~\cite{sadeghian19,jiachen19,kosaraju19,dendorfer20,kothari21}.
The core of this work is context-free methods: learning to predict future trajectories based on observed motion actions. We investigate self-supervised approaches that include meaningful behavioral-related information in the forecaster.

Recurrent Neural Networks (RNNs) are broadly used networks' structures for solving sequential data problems~\cite{ilya14, wang18}. 
Due to their prominent behavior for modeling dynamics, they have also been used in the context of motion analysis \cite{morton17,becker18}. There are two variants of RNNs: Long-Short Term Memory (LSTM)~\cite{hochreiter97}, and the lighter version Gated Recurrent Unit (GRU)~\cite{Cho14}.
In the scope of our work, these two mechanisms play a relevant role in extracting temporal features. Those features can then be processed 
for forecasting or analytics purposes.

Standalone LSTMs may be a suitable choice for modeling the average behavior. Indeed, it is a decent design choice for most of the 
current benchmarks as the diversity of the presented data. Conversely, there are real-world problems where one aims to model the distribution of
the data as a whole (e.g., surveillance systems where the average behavior is not suitable). In this line, Generative Adversarial Networks~\cite{ian14} took over the motion forecasting scene since it provides a group of plausible predictions.  

For the image generation task, GANs have been demonstrating interesting results~\cite{radford16, heusel17, karras19,karras20},
whereas it seems still open the usage of these frameworks in the Motion Analysis domain. Still in the Computer Vision domain, one of the current research lines 
is the usage of GANs in unsupervised tasks such as clustering~\cite{jie18, mukherjee19}. Analogously,~\cite{karras19} generates images conditioned on 
different cluster classes. In this work, the clustering space comprises features learned by the discriminator. Despite this, the use of the features from the 
discriminator in downstream tasks is still an open question. 

In the Motion Analysis domain, traditional clustering methods
have been researched~\cite{bian18,ahmed19} yet used in a standalone fashion. 
To the best of our knowledge, we propose the first framework that clusters features yielded by 
Generative models and provides directions for improving the training procedure. These training guidelines foster the learning of the least dominant modes in the data, which 
means that it helps to attenuate the mode collapse problem, a well-researched matter by the Computer Vision community~\cite{akash17, mao19}.

\section{METHODOLOGY}

We aim to model broader context-free motion patterns. Current state-of-the-art approaches aim to outperform benchmark results according 
to overall metrics such as Average Displacement Error (ADE) and Final Displacement Error (FDE). Instead, our goal is to learn better the diversity
of trajectories. In this section, we first formulate the problem of context-free 
motion prediction. Secondly, inspired by~\cite{liu20}, we propose our main framework to learn the different modes from the data. Finally, we present
different training settings propelling a more balanced forecaster, i.e., less biased to the most dominant behavior and capable of spanning more modes.

\subsection{Problem Definition}

This work formalizes a dual-objective system. The first goal is to get insights from different behaviors --- modes --- based on motion data. 
These modes, $\boldsymbol{m} = (m_1, ..., m_{k})$, are one-hot encoded variables representing $k$ clusters formed via discriminative features captured within a GAN. To do so,
the inputs of the generative mode learner are a batched $\boldsymbol{\rm X} = (\boldsymbol{x_1}, ..., \boldsymbol{x_n})$ representing 
$n$ different trajectories and the respective unsupervised modes. Each trajectory, $\boldsymbol{x_i}$, is composed of $t$ 2D directional vectors (in $XY$\nobreakdash-plane) from the observed part of 
the trajectory. Analogously, the goal is to generate $\boldsymbol{\rm Y} = (\boldsymbol{y_1}, ..., \boldsymbol{y_n})$ comprising $h$ 2D incremental steps. The second objective stands for improving the capacity to generate various modes of a Vanilla GAN forecaster, where the input is uniquely $\boldsymbol{\rm X}$, by using insights from the learned modes, $\boldsymbol{m}$.

\subsection{Method}

Our system (Fig.\ref{Fig1:framework}) is a two-step approach. First, we learn the different
modes of input data through the self-conditioned GAN. In doing so, we obtain privileged information
about the different clusters. Since those clusters derive from entire trajectories, 
we can know which is the most representative or yields the worst results. Second, by using this 
information, we design training settings that encourage learning more diverse modes.

\subsubsection{self-conditioned GAN}

The system responsible for learning the different modes relies on a GAN paradigm. It adversarially trains a generator $G$ and a discriminator $D$. 
The former learns to produce fake samples ($\boldsymbol{{\hat{Y}}}$) similar as much as possible to the real training data, $p(x)$, while the latter distinguishes real and generated samples. 
To do so, $G$ takes as an input a latent variable ($\boldsymbol{z}$) sampled from a standard Gaussian distribution, $p(z)$. In the presented problem, besides the latent variable, $G$ is also conditioned
on the observed track ($\boldsymbol{\rm X}$). Regarding the discriminator, it takes both real ($\boldsymbol{R} = \boldsymbol{\rm X}~\oplus~\boldsymbol{Y}$) and generated samples 
($\boldsymbol{\hat{F}} = \boldsymbol{\rm X}~\oplus~\boldsymbol{{\hat{Y}}}$), mapping them to the respective likelihood ($\boldsymbol{s}$) of coming from the real distribution. From the networks' training point of view, the discriminator 
provides the training signal to the generator by evaluating how different the generated samples are from the real ones. The final objective is that both $D$ and $G$  
reach the Nash equilibrium, where each network cannot decrease its own cost without changing the other's network parameters~\cite{fedus18}. Fundamentally, both networks play a two-player mini-max game by optimizing a value function, $V$:

\begin{equation}
    \begin{aligned}
        & \min_{G} \max_{D} V(G, D)  =\\
        & = \mathbb{E} [\log D(x \oplus y)] +  \mathbb{E}[\log(1 - D(G(x, z)))],
    \end{aligned}
\end{equation}
 where $x~\oplus~y~\sim~p(x~\oplus~y)$, $x~\sim~p(x)$, and $z~\sim~p(z)$.

Although GANs enable the spanning of different modes from the original distribution, if the data set is biased to a dominant subspace of the 
data distribution, GANs suffer from mode collapse. It means that the generator would focus intently on the dominant behaviors and do not span the remaining modes.
Conditional GANs~\cite{mehdi14} came to overcome some of the barriers imposed by this problem. By adding an explicit condition such as a label class, one has more control 
under the modes spanned by the generator. In the presented problem, there is already a condition, the observed track ($\boldsymbol{\rm X}$). However, due to the complexity of the input data, 
the intuition is that such a condition is not explicit enough to drive our generator in a way that can fully span the different motion profiles. The hypothesis is that since part of the discriminator's input 
is real samples, clustering its features can lead to a fair approximation of the different real modes. This reasoning is similar to the one described in~\cite{liu20}, but we converted it to the 
motion prediction problem. 

\begin{figure}[ht]
    \centering
    \tikzset{myLine1/.style = {->, thick, >=stealth},
    mynarrownodes/.style = {node distance=0.75cm and 0.3cm},
    ball/.style={
        ellipse,
        minimum width=3cm,
        minimum height=1.5cm,
        draw
    },
    }
    
    \begin{tikzpicture}
        \tikzstyle{bigbox} = [draw=black!50, thick, fill=gray!50, rounded corners, rectangle]
        \tikzstyle{box} = [minimum size=0.2cm, rounded corners,rectangle split,rectangle split parts=4, inner sep=0.2ex, fill=gray!10]
        
        \node[inner sep=0, anchor=center] (input_disc_fake) at (0,0) {$\boldsymbol{\hat{F}}$};
        \node[align=center,below=0.50mm of input_disc_fake](or) {or};
        \node[align=center,below=0.05mm of or](input_disc_real) {$\boldsymbol{\rm R}$};

        \matrix[right=1.20cm of or, row sep=0mm, column sep=7mm, inner sep=2mm, bigbox, every node/.style=box](disc_box) {
            \node(encoder_box){\textbf{Encoder} \vphantom{$\vcenter{\vspace{1.5em}}$} \nodepart{second} LSTM  \nodepart{third} or \nodepart{fourth} MLP}; 
            &  \node(classifier_box){\textbf{Classifier} \vphantom{$\vcenter{\vspace{1.5em}}$} \nodepart{second} \nodepart{third} MLP \nodepart{fourth}}; \\
        };
        \node[align=center, above=0.25mm of disc_box](disc_string) {DISCRIMINATOR ($D$)};
        \node[align=center, right=7cm of or](scores) {$\boldsymbol{s}$};

        \draw[myLine1] (or.east) to[out=0, in=180] (disc_box.west);
        \draw[myLine1] (disc_box.east) to[out=0, in=180] (scores.west);
        \draw[myLine1] (encoder_box) -- (classifier_box) node [sloped,midway](M){};

        \node[ball, align=center,below=0.5cm of disc_box](clustering) {};
        
        \node[circle,draw,fill=gray!25, minimum size=0.1cm] at (4.65,-2.72) (m1) {};
        \node[ellipse,rotate=35,draw,fill=gray!25,minimum width=1.2cm,minimum height=0.5cm,above left=0.30cm of m1](m2) {};
        \node[ellipse,rotate=80,draw,fill=gray!25,minimum width=0.3cm,minimum height=1.0cm,left=0.30cm of m2](m3) {};
        \node[circle,draw,fill=gray!25, minimum size=0.40cm, above=0.2cm of m3](m4) {};
        \node[ellipse,draw,fill=gray!25,minimum width=0.2cm,minimum height=0.60cm,right=0.2cm of m4](m5) {};
        \node[ellipse,rotate=60,draw,fill=gray!25,minimum width=0.23cm,minimum height=0.64cm,below right=0.3cm of m3](m6) {};
        
        \draw[myLine1] (M) -- (clustering);
        
        \node[align=center,below=4.4cm of or, draw, rectangle, fill=white](input_gen){$\boldsymbol{\rm X}\oplus\boldsymbol{m}$};
        \matrix[right=0.85cm of input_gen, row sep=0mm, column sep=10mm, inner sep=2mm, bigbox, every node/.style=box](gen_box) {
            \node(encoder_box2){\textbf{Encoder} \vphantom{$\vcenter{\vspace{1.5em}}$} \nodepart{second}  \nodepart{third} LSTM \nodepart{fourth} }; 
            &  \node(decoder_box){\textbf{Decoder} \vphantom{$\vcenter{\vspace{1.5em}}$} \nodepart{second} \nodepart{third} LSTM \nodepart{fourth}}; \\
        };
        \node[align=center, above=0.25mm of gen_box](gen_string) {GENERATOR ($G$)};
        \node[align=center, right=1.2cm of gen_box](pred) {$\boldsymbol{{\hat{Y}}}$};
        \node[align=center, left=0.25mm of clustering](cluster_str) {Clustering};
        \draw[myLine1] (input_gen.east) to[out=0, in=180] (gen_box.west);
        \draw[myLine1] (gen_box.east) to[out=0, in=180] (pred.west);
        \node[align=center, right=2.30mm of encoder_box2](concat_noise) {$\oplus$};
        \node[align=center, below=3.0mm of concat_noise](z) {$\boldsymbol{z}$};
        \draw[myLine1] (encoder_box2.east) to[out=0, in=180] (concat_noise.west);
        \draw[myLine1] (concat_noise.east) to[out=0, in=180] (decoder_box.west);
        \draw[myLine1] (z.north) to[out=90, in=270] (concat_noise.south);
        \draw[myLine1] (m3.west) to[out=-180,   in=90] (input_gen.40);

    \end{tikzpicture}
    \caption{Self-conditioned GAN for identifying meaningful clusters, which represent different modes ($\boldsymbol{m}$) of the input trajectory data. It comprises two branches: the discriminator ($D$) and the generator ($G$). 
    During the former's training (upper part), the inputs are fake or real samples, $\boldsymbol{{\hat{F}}}$ and $\boldsymbol{R}$, respectively. 
    Then, the encoder extracts features from those inputs. These features become semantically meaningful during training, so they are reclustered from time to time. 
    Finally, a classifier produces the likelihood, $\boldsymbol{s}$, of the respective input being a sample from the real distribution. The second branch (lower part) depicts the generator's training. It takes as input the 
    observed trajectories, $\boldsymbol{X}$, concatenated with the respective modes, $\boldsymbol{m}$. Before this, the clustering algorithm provides the respective modes based on the discriminator's encoder
    features from the entire and real trajectory, $\boldsymbol{R}$. Thereafter, the generator's encoder extracts features from the input. Finally, these features plus the latent variable, $\boldsymbol{z}$, are decoded, yielding
    the predicted trajectory, $\boldsymbol{{\hat{Y}}}$.}
    \label{Fig2:self-conditioned GAN}
\end{figure}
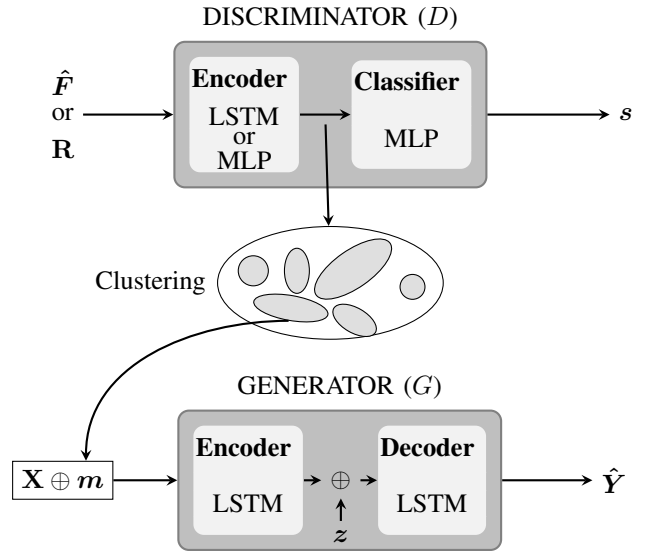

To the best of our knowledge, this is the first work where a self-conditioned GAN is redesigned, aiming to capture different modes from trajectories data. Fig.~\ref{Fig2:self-conditioned GAN} depicts
the overview of the described methodology for modes recognition. It is worth mentioning that there are two possible encoders in the discriminator: an LSTM  and a MLP. Discriminator's encoder consisting of LSTMs was a 
design choice in~\cite{gupta18, sadeghian19, kosaraju19}. Conversely, in~\cite{kothari19}, the authors claimed a more stable training by using an MLP, so we opted for
the latter. For a more detailed description of the self-conditioned GAN training, the reader is pointed to~\cite{liu20}.

\subsubsection{Soft Assumptions and Training Settings}

After learning the different modes of the data represented by feature-based clusters, our goal is to recover better these modes from the input data. 
Therefore, we (softly) assume that those clusters are grouping similar  motion profiles. In this way, clusters representing 
more challenging trajectories, according to the prediction error yielded by the self-conditioned GAN, are the ones corresponding to the modes less prone to be recovered. 

Hence,  we embed this information into three vanilla GAN training settings. First, we propose the usage of a \emph{weighted generator loss} (wL2) based on the distribution of the clustering space
and the intra-cluster results obtained by the self-conditioned GAN. 
It forces the generator to emphasize the learning of samples from more challenging subspaces (representing modes more difficult 
to recover). Therefore, the novel generator's loss, initially given
by the sum of the adversarial loss and the \emph{L2 loss}, penalizes the latter term, as follows:

\begin{equation}
    \Lambda_i = \lambda_{ADE}~\frac{ADE_i}{ADE_{\text{max}}}+\lambda_{FDE}~\frac{FDE_i}{FDE_{\text{max}}}+\lambda_{D}~\frac{\#_i}{\#_{\text{T}}},
\label{eq2:weights}
\end{equation}
 where $\Lambda_i$ is the weight applied to the trajectories in cluster $i$; $ADE$ and $FDE$ are the Average Displacement Error
and the Final Displacement Error, respectively, yielded by the modes recognizer; $\#_i$ is the number of samples in cluster $i$ and $\#_{\text{T}}$ is the total 
number of samples in the clustering space. This last term mitigates outliers in the input data. Finally, $\lambda$
stands for weights applied to each of the respective terms.   

Additionally, we deployed a \emph{weighted batch sampler} (wB) based on a multinomial distribution where the weights of the respective 
classes are provided by (\ref{eq2:weights}). Lastly, the third training setting (wL2+wB) joins the two previously described settings.

\section{EXPERIMENTS}




In this section, we show the results of our methods in two different data sets: TH\"{O}R~\cite{rudenko20} and Argoverse~\cite{ming19}. Firstly, we present quantitative results on the 
trajectories generation. Secondly, we analyze qualitatively generated tracks and present insights on the self-conditioned GAN.

\subsection{Description of data sets}

The Argoverse data set consists of trajectories collected in road environments from three different moving agents: 
autonomous vehicles ({\em av}), regular vehicles ({\em agents}), and other road agents ({\em others}). 
The trajectories consist of observed tracks of 20-time steps (\SI{2}{s}) plus 30-time steps (\SI{3}{s}) of prediction.
Similar to~\cite{chandra20}, we randomly sampled 5726, 2100, and 1678 trajectories for training, validation, and testing sets,  
respectively. However, to study the learning of the least representative supervised classes, we set a constraint on the creation of the 
training set: 2600 trajectories from {\em av}, 2600 from {\em agents}, and 526 from {\em others} (factor of $\approx 5$ times).

The TH\"{O}R data set contains trajectories of humans in an industrial-like environment. It includes three experiments where people with different roles perform several task-driven tracks. 
The roles include 5 or 6 {\em visitors}, 2 {\em workers}, and 1 {\em inspector}. Inspired by benchmarks~\cite{kothari21}, we split each trajectory into tracks of 8-time steps (\SI{3.2}{s}) for observation plus 12-time steps (\SI{4.8}{s}) for 
prediction. In order to use the data in our experiments, and based on~\cite{rudenko21}, we applied the following preprocessing steps: (1) downsample the signal to \SI{400}{ms}, (2) linearly interpolate 
missing detections and (3) smooth the resulting trajectories with a moving average filter (\SI{800}{ms} window). 
For this, we developed the \emph{pythor-tools} framework~\cite{pythor}.

Finally, both data sets provide supervised labels (types of moving agents and human roles), which we use to evaluate the proposed methods (i.e., intra-supervised label performance).

\subsection{Quantitative results}

In the first experiment, we show the intra-supervised label performance of our approaches. Following previous works~\cite{gupta18, sadeghian19, kosaraju19, jiachen19, dendorfer20,kothari21}
ADE calculates the Root Mean Squared Error (RMSE) between the ground truth prediction and the estimated track. FDE calculates the distance between the $(x,y)$ positions 
in the last time step between the real and the predicted trajectory. Throughout this section, we report generative models' scores as an average of 5 runs. "wL2" and "wB" abbreviate the weighted loss and the weighted batch sampler approaches, respectively. Finally, bold scores mean the best method's results.

We compare our proposed training settings with two context-free approaches: a simple LSTM followed by an MLP~\cite{becker18}, and a Vanilla GAN, inspired by context-free methods provided by
Trajnet++~\cite{kothari21}. In contrast to other previous works~\cite{gupta18, sadeghian19, kosaraju19, jiachen19, dendorfer20,kothari21}, the results presented 
throughout this section are based on one predicted trajectory from the generative models. This implies a fairer comparison between generative and deterministic approaches. The self-conditioned GAN is grounded on a clustering method to capture akin 
discriminative features. Thus, it relies on a clustering algorithm and, therefore, on the number of clusters. We found both optimal settings through a grid search on the validation sets: \emph{k-Means}~\cite{jin10} approach with 13 clusters for the TH\"{O}R and 19 for the Argoverse data sets.

Table~\ref{tab:overall_results} presents our forecasting results in comparison with other methods, where underlined supervised classes are the least predominant in the training set. According to the ADE and FDE metrics in the least representative profiles, 
our approach based on Vanilla GAN+wB and Vanilla GAN+wL2+wB outperformed previous non-context approaches in both data sets. Therefore, the proposed training settings drive the generator to learn the most challenging unsupervised subspaces, which improves the forecasting results for the least representative supervised labels. Furthermore, in the TH\"{O}R data set, the difference between the results for the {\em inspector} is not as significant as for the {\em workers} role. This might be due to the data set designing since {\em workers} had to carry materials (e.g., boxes), whereas the {\em inspector} and {\em visitors} have much more similar behaviors~\cite{rudenko20}.

\begin{table}[t]
  \centering
  \caption{Intra-classes ADE/FDE metrics (in meters) in the test sets.}
  \label{tab:overall_results}
  \resizebox{\linewidth}{!}{%
  \begin{tabular}{@{}cc|cc|ccc@{}}
  \multicolumn{2}{l|}{} &
    \multicolumn{2}{c|}{\textbf{Baselines}} &
    \multicolumn{3}{c}{\textbf{Ours}} \\ \midrule
  Data set &
    \begin{tabular}[c]{@{}c@{}}Labels \\ (\# samples)\end{tabular} &
    \begin{tabular}[c]{@{}c@{}}LSTM\\\cite{becker18}\end{tabular} &
    \begin{tabular}[c]{@{}c@{}}CF VAN \\GAN~\cite{kothari21}\end{tabular} &
      \begin{tabular}[c]{@{}c@{}}CF VAN \\ GAN + wL2\end{tabular} &
      \begin{tabular}[c]{@{}c@{}}CF VAN \\ GAN + wB\end{tabular} &
      \begin{tabular}[c]{@{}c@{}}CF VAN GAN\\ +wL2 + wB\end{tabular} \\ \midrule
  \multirow{6}{*}{TH\"{O}R} &
    \begin{tabular}[c]{@{}c@{}}\underline{{\em workers}} \\ (413)\end{tabular} &
    \begin{tabular}[c]{@{}c@{}}0.695\\ 1.064\end{tabular} &
    \begin{tabular}[c]{@{}c@{}}$0.642\pm0.006$\\$1.033\pm0.005$\end{tabular} &
    \begin{tabular}[c]{@{}c@{}}$0.629\pm0.005$\\$1.009\pm0.014$\end{tabular} &
    \begin{tabular}[c]{@{}c@{}}$0.644\pm0.012$\\$1.044\pm0.028$\end{tabular} &
    \textbf{\begin{tabular}[c]{@{}c@{}}$\mathbf{0.625\pm0.009}$\\$\mathbf{1.006\pm0.019}$\end{tabular}} \\ \cmidrule(l){2-7} 
   &
    \begin{tabular}[c]{@{}c@{}}{\em visitors} \\ (1379)\end{tabular} &
    \begin{tabular}[c]{@{}c@{}}0.664\\ 1.139\end{tabular} &
    \begin{tabular}[c]{@{}c@{}}$0.660\pm0.001$\\$\mathbf{1.105\pm0.090}$\end{tabular} &
    \begin{tabular}[c]{@{}c@{}}$\mathbf{0.657\pm0.003}$\\$1.107\pm0.007$\end{tabular} &
    \begin{tabular}[c]{@{}c@{}}$0.668\pm0.005$\\$1.124\pm0.018$\end{tabular} &
    \begin{tabular}[c]{@{}c@{}}$0.657\pm0.003$\\$1.113\pm0.013$\end{tabular} \\ \cmidrule(l){2-7} 
   &
    \begin{tabular}[c]{@{}c@{}}\underline{{\em inspector}} \\ (260)\end{tabular} &
    \begin{tabular}[c]{@{}c@{}}0.796\\ 1.582\end{tabular} &
    \begin{tabular}[c]{@{}c@{}}$0.735\pm0.007$\\$1.474\pm0.019$\end{tabular} &
    \begin{tabular}[c]{@{}c@{}}$0.736\pm0.008$\\$\mathbf{1.473\pm0.013}$\end{tabular} &
    \textbf{\begin{tabular}[c]{@{}c@{}}$\mathbf{0.729\pm0.013}$\\$1.479\pm0.049$\end{tabular}} &
    \begin{tabular}[c]{@{}c@{}}$0.734\pm0.003$\\$1.476\pm0.015$\end{tabular} \\ \midrule
  
    \multirow{6}{*}{Argoverse} &
    \begin{tabular}[c]{@{}c@{}}\underline{{\em others}} \\ (526)\end{tabular} &
    \begin{tabular}[c]{@{}c@{}}1.864\\ 3.029\end{tabular} &
    \begin{tabular}[c]{@{}c@{}}$1.815\pm0.031$\\$2.969\pm0.034$\end{tabular} &
    \begin{tabular}[c]{@{}c@{}}$1.799\pm0.007$\\$2.944\pm0.022$\end{tabular} &
    \textbf{\begin{tabular}[c]{@{}c@{}}$\mathbf{1.789\pm0.012}$\\$2.927\pm0.020$\end{tabular}} &
    \begin{tabular}[c]{@{}c@{}}$1.801\pm0.027$\\ $\mathbf{2.919\pm0.032}$\end{tabular} \\ \cmidrule(l){2-7} 
   &
    \begin{tabular}[c]{@{}c@{}}{\em av} \\ (2600)\end{tabular} &
    \begin{tabular}[c]{@{}c@{}}1.512\\ 2.278\end{tabular} &
    \begin{tabular}[c]{@{}c@{}}$\mathbf{1.467\pm0.007}$\\$\mathbf{2.269\pm0.023}$\end{tabular} &
    \begin{tabular}[c]{@{}c@{}}$1.482\pm0.009$\\$2.292\pm0.010$\end{tabular} &
    \begin{tabular}[c]{@{}c@{}}$1.480\pm0.003$\\$2.282\pm0.006$\end{tabular} &
    \begin{tabular}[c]{@{}c@{}}$1.493\pm0.010$\\$2.298\pm0.028$\end{tabular} \\ \cmidrule(l){2-7} 
   &
    \begin{tabular}[c]{@{}c@{}}{\em agent} \\ (2600)\end{tabular} &
    \begin{tabular}[c]{@{}c@{}}2.371\\ 4.690\end{tabular} &
    \begin{tabular}[c]{@{}c@{}}$\mathbf{2.349\pm0.012}$\\$\mathbf{4.654\pm0.016}$\end{tabular} &
    \begin{tabular}[c]{@{}c@{}}$2.362\pm0.013$\\$4.700\pm0.029$\end{tabular} &
    \begin{tabular}[c]{@{}c@{}}$2.368\pm0.020$\\$4.721\pm0.044$\end{tabular} &
    \begin{tabular}[c]{@{}c@{}}$2.371\pm0.012$\\$4.724\pm0.027$\end{tabular} \\ \bottomrule
  \end{tabular}%
  }
  \text{}
  \end{table}

  Table~\ref{tab:intra-cluster} reports results in two groups of trajectories according to the self-conditioned GAN's clustering space. We show the results for the most challenging and smallest cluster (where the self-conditioned GAN yielded the worst intra-cluster results), clusters 9 and 10 for TH\"{O}R and Argoverse, respectively; and the most dominant (i.e., the one with  the largest number of samples, clusters 0 and 18 for TH\"{O}R and Argoverse, respectively). As expected, this Table shows that there is a 
  correlation between the number of trajectories of each cluster and the metrics obtained by each method. That is, clusters with fewer trajectories present worse ADE/FDE scores, while dominant clusters present better results.
  Any, our proposed training settings outperformed the baselines in the most challenging clusters validating the described \emph{soft-assumptions} and presenting comparable results
  in the most dominant group of trajectories. 

\begin{table}[t]
  \centering
  \caption{ADE/FDE metrics (in meters) for 2 clusters of the test set.}
  \label{tab:intra-cluster}
  \resizebox{\linewidth}{!}{%
  \begin{tabular}{@{}cc|cc|ccc@{}}
  \multicolumn{2}{c|}{} &
    \multicolumn{2}{c|}{\textbf{Baselines}} &
    \multicolumn{3}{c}{\textbf{Ours}} \\ \midrule
  Data set &
    \begin{tabular}[c]{@{}c@{}}Cluster ID\\ (\#  samples)\end{tabular} &
    \begin{tabular}[c]{@{}c@{}}LSTM\\\cite{becker18}\end{tabular} &
    \begin{tabular}[c]{@{}c@{}}CF VAN \\GAN~\cite{kothari21}\end{tabular} &
      \begin{tabular}[c]{@{}c@{}}CF VAN \\ GAN + wL2\end{tabular} &
      \begin{tabular}[c]{@{}c@{}}CF VAN \\ GAN + wB\end{tabular} &
      \begin{tabular}[c]{@{}c@{}}CF VAN GAN\\ +wL2 + wB\end{tabular} \\ \midrule
    \multirow{4}{*}{TH\"{O}R} &
      \begin{tabular}[c]{@{}c@{}}9 \\ (23)\end{tabular} &
      \begin{tabular}[c]{@{}c@{}}1.203\\ 2.456\end{tabular} &
      \begin{tabular}[c]{@{}c@{}}$1.120\pm0.025$\\$2.758\pm0.082$\end{tabular} &
      \begin{tabular}[c]{@{}c@{}}$1.054\pm0.048$\\$2.505\pm0.134$\end{tabular} &
      \begin{tabular}[c]{@{}c@{}}$1.124\pm0.038$\\$2.811\pm0.136$\end{tabular} &
      \textbf{\begin{tabular}[c]{@{}c@{}}$\mathbf{1.039\pm0.055}$\\$\mathbf{2.505\pm0.105}$\end{tabular}} \\ \cmidrule(l){2-7} 
     &
      \begin{tabular}[c]{@{}c@{}}0 \\ (1003)\end{tabular} &
      \begin{tabular}[c]{@{}c@{}}0.325\\ 0.447\end{tabular} &
      \begin{tabular}[c]{@{}c@{}}$\mathbf{0.311\pm0.003}$\\$\mathbf{0.403\pm0.010}$\end{tabular} &
      \begin{tabular}[c]{@{}c@{}}$0.321\pm0.004$\\$0.419\pm0.015$\end{tabular} &
      \begin{tabular}[c]{@{}c@{}}$0.317\pm0.007$\\$0.416\pm0.021$\end{tabular} &
      \begin{tabular}[c]{@{}c@{}}$0.315\pm0.002$\\$0.424\pm0.017$\end{tabular} \\ \midrule
  \multirow{4}{*}{Argoverse} &
    \begin{tabular}[c]{@{}c@{}}10 \\ (16)\end{tabular} &
    \begin{tabular}[c]{@{}c@{}}7.394\\ 19.075\end{tabular} &
    \begin{tabular}[c]{@{}c@{}}$7.184\pm0.178$\\$18.402\pm0.415$\end{tabular} &
    \begin{tabular}[c]{@{}c@{}}$7.105\pm0.123$\\$18.233\pm0.297$\end{tabular} &
    \begin{tabular}[c]{@{}c@{}}$7.122\pm0.055$\\$18.276\pm0.113$\end{tabular} &
    \textbf{\begin{tabular}[c]{@{}c@{}}$\mathbf{7.047\pm0.088}$\\$\mathbf{18.128\pm0.194}$\end{tabular}} \\ \cmidrule(l){2-7} 
   &
    \begin{tabular}[c]{@{}c@{}}18 \\ (1542)\end{tabular} &
    \begin{tabular}[c]{@{}c@{}}0.912\\ 1.148\end{tabular} &
    \begin{tabular}[c]{@{}c@{}}$0.809\pm0.016$\\$1.100\pm0.017$\end{tabular} &
    \begin{tabular}[c]{@{}c@{}}$0.807\pm0.010$\\$1.088\pm0.027$\end{tabular} &
    \begin{tabular}[c]{@{}c@{}}$0.805\pm0.007$\\$1.079\pm0.012$\end{tabular} &
    \begin{tabular}[c]{@{}c@{}}$\mathbf{0.795\pm0.008}$\\$\mathbf{1.055\pm0.030}$\end{tabular} \\ \bottomrule
  \end{tabular}%
  }
  \end{table}

Finally, Table~\ref{tab:overall} shows the overall results obtained in both data sets. While in the TH\"{O}R data set, Vanilla GAN+wL2 could improve the average results in both metrics, 
in the Argoverse, as we forced the number of samples of the least representative class ({\em others}) to be very small, we could not improve the 
average scores. Though, we could improve the performance of the least dominant profiles in both data sets.

\begin{table}[t]
    \centering
    \caption{ADE/FDE metrics (in meters) in the test sets.}
    \label{tab:overall}
    \resizebox{\linewidth}{!}{%
    \begin{tabular}{@{}c|cc|ccc@{}}
     &
      \multicolumn{2}{c|}{\textbf{Baselines}} &
      \multicolumn{3}{c}{\textbf{Ours}} \\ \midrule
    Data set &
      \begin{tabular}[c]{@{}c@{}}LSTM\\\cite{becker18}\end{tabular} &
      \begin{tabular}[c]{@{}c@{}}CF VAN \\GAN~\cite{kothari21}\end{tabular} &
      \begin{tabular}[c]{@{}c@{}}CF VAN \\ GAN + wL2\end{tabular} &
      \begin{tabular}[c]{@{}c@{}}CF VAN \\ GAN + wB\end{tabular} &
      \begin{tabular}[c]{@{}c@{}}CF VAN GAN\\ +wL2 + wB\end{tabular} \\ \midrule
    TH\"{O}R &
      \begin{tabular}[c]{@{}c@{}}$0.685$\\$1.163$\end{tabular} &
      \begin{tabular}[c]{@{}c@{}}$0.663\pm0.002$\\$1.123\pm0.019$\end{tabular} &
      \begin{tabular}[c]{@{}c@{}}$\mathbf{0.658\pm0.003}$\\$\mathbf{1.119\pm0.009}$\end{tabular} &
      \begin{tabular}[c]{@{}c@{}}$0.668\pm0.003$\\$\mathbf{1.119\pm0.009}$\end{tabular} &
      \begin{tabular}[c]{@{}c@{}}$\mathbf{0.658\pm0.002}$\\$1.122\pm0.013$\end{tabular} \\ \midrule
    Argoverse &
      \begin{tabular}[c]{@{}c@{}}$1.948$\\$3.330$\end{tabular} &
      \begin{tabular}[c]{@{}c@{}}$\mathbf{1.912\pm0.012}$\\$\mathbf{3.298\pm0.014}$\end{tabular} &
      \begin{tabular}[c]{@{}c@{}}$1.915\pm0.012$\\$3.307\pm0.017$\end{tabular} &
      \begin{tabular}[c]{@{}c@{}}$1.915\pm0.009$\\$3.310\pm0.018$\end{tabular} &
      \begin{tabular}[c]{@{}c@{}}$1.923\pm0.010$\\$3.307\pm0.014$\end{tabular} \\ \bottomrule
    \end{tabular}%
    }
\end{table}

\subsection{Self-conditioned GAN analysis and discussion}

A key component of our proposed approach is the self-conditioned GAN, which provides the signals for the proposed training settings. 
One of the drawbacks of the clustering procedure is the difficulty of understanding what type of trajectory each cluster may represent. 
Assuming that each resulting cluster represents a group of trajectories sharing common patterns, we may use the cluster id as a label for each group. 
Then, the corresponding cluster id is the condition of the self-conditioned GAN generator. 
In order to validate this hypothesis, we compare our approach to a conditional GAN. 
In our case we use the resulting {\em clusters} classes as conditions, and we compare with the case where the conditions correspond 
to the different profiles in the data sets: {\em av}, {\em agents}, and {\em others} in Argoverse; and {\em visitors}, {\em workers}, and {\em inspector} in TH\"{O}R. 
These two models are named {\em ideal} as in real scenarios we can not obtain these conditions uniquely by processing the observed part of the trajectory. 
Table~\ref{tab:overall_ideal} shows the comparison between cGAN and the presented self-conditioned GAN. The ADE/FDE metrics for the self-conditioned GAN can be seen as a lower bound for those metrics and support our idea
of using clustering to recognize different motion-based modes.

\begin{table}[t]
  \centering
  \caption{ADE/FDE metrics (in meters) in the test sets for {\em ideal} models. These can be seen as lower bounds.}
  \label{tab:overall_ideal}
  \begin{tabular}{@{}l|ll@{}}
  Data set & cGAN                                                                & Ours                                                              \\ \midrule
  TH\"{O}R      & \begin{tabular}[c]{@{}l@{}}$0.657\pm0.003$\\$1.114\pm0.011$\end{tabular} & \begin{tabular}[c]{@{}l@{}}$\mathbf{0.591\pm0.014}$\\$\mathbf{0.937\pm0.022}$\end{tabular} \\ \midrule
  Argoverse     & \begin{tabular}[c]{@{}c@{}}$1.927\pm0.016$\\$3.323\pm0.030$\end{tabular} & \begin{tabular}[c]{@{}c@{}}$\mathbf{1.785\pm0.014}$\\$\mathbf{2.887\pm0.038}$\end{tabular} \\ \bottomrule
  \end{tabular}%
 \end{table}

\subsection{Qualitative Analysis}
\begin{figure}[t]
   \centering
        \includegraphics[width=0.48\columnwidth]{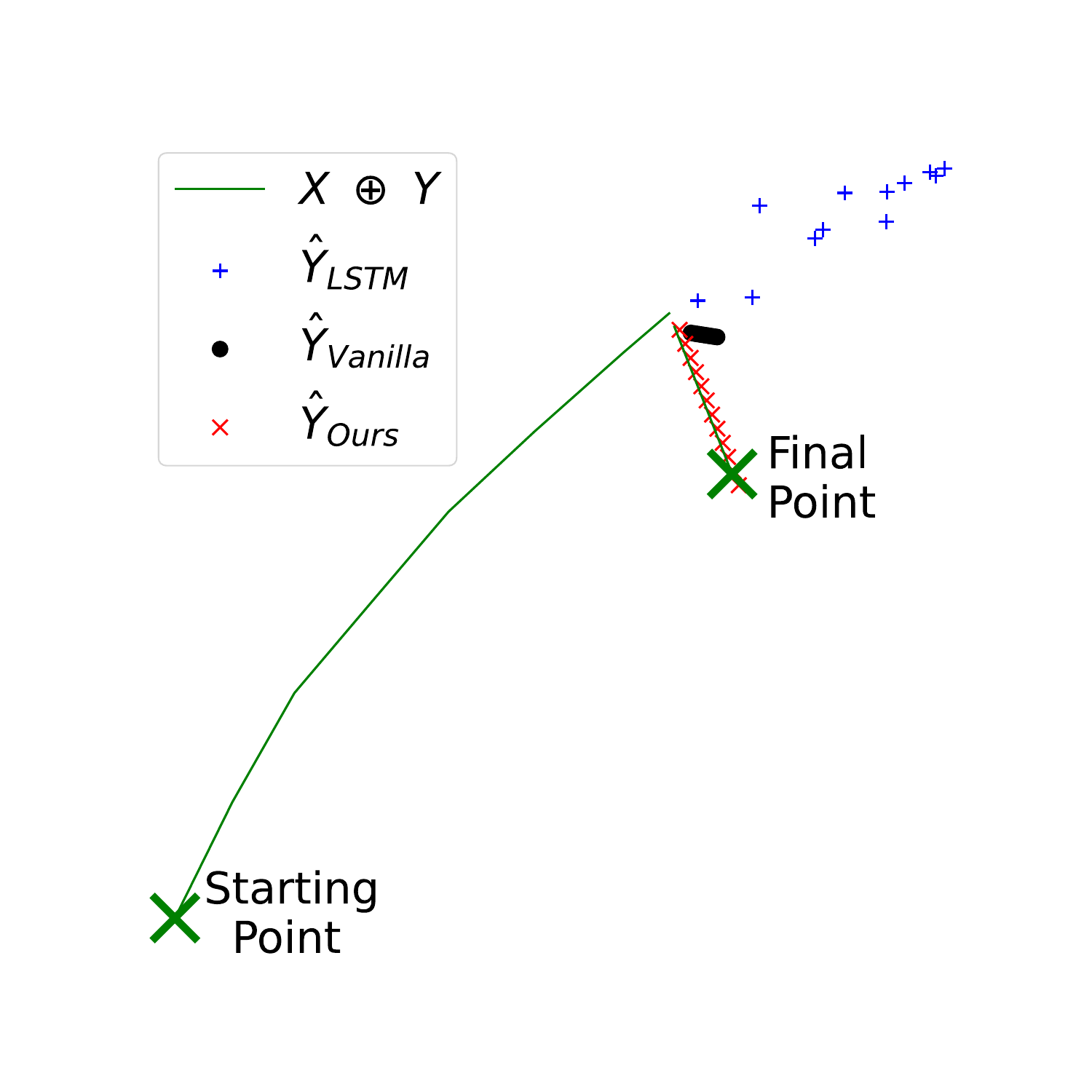}
        \includegraphics[width=0.48\columnwidth]{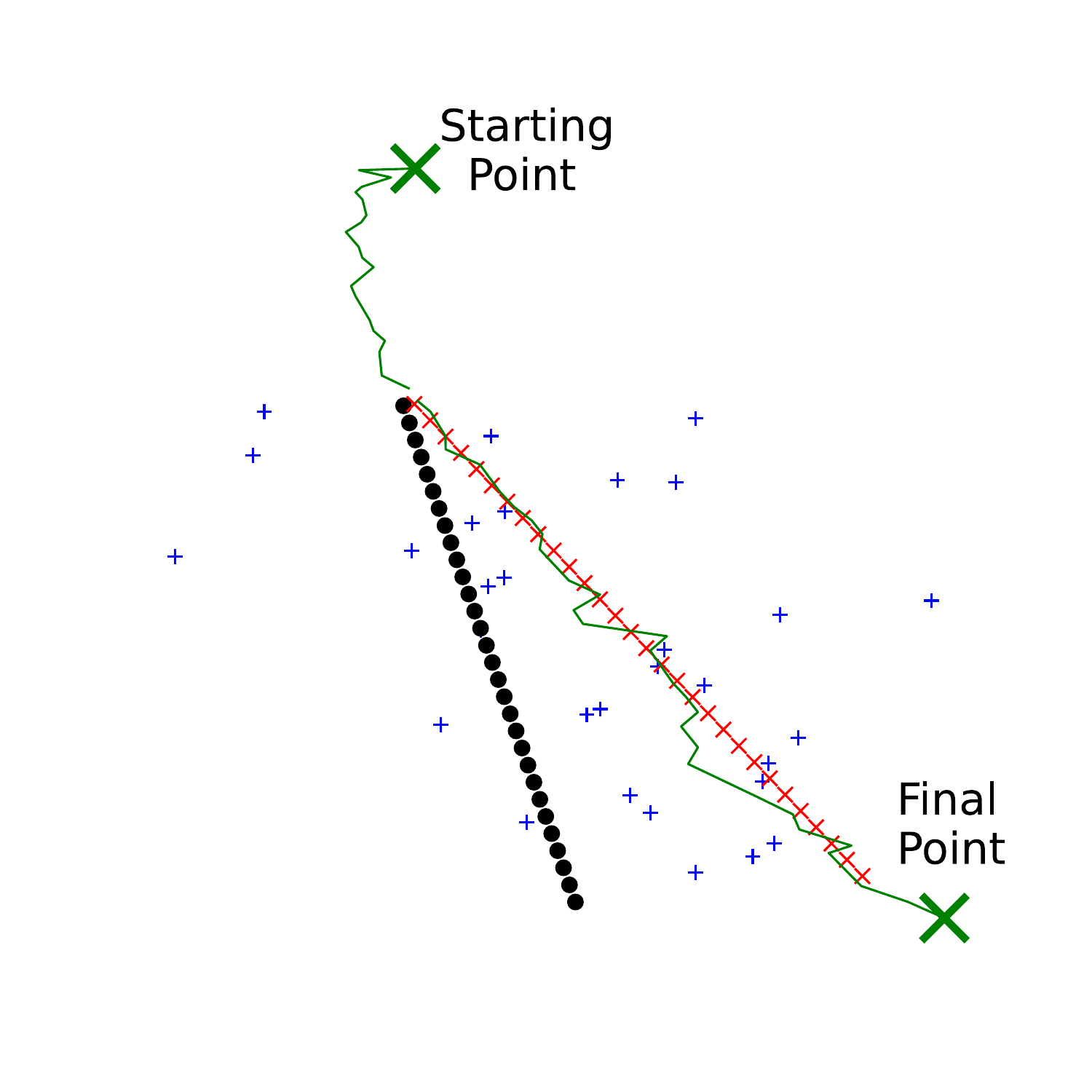}                      
    \caption{Example of trajectory forecasting in the TH\"{O}R (left) and the Argoverse (right) data sets for complex trajectories, being $X$ the observed track and $Y$ and $\hat{Y}$ the ground truth and the predictions, respectively. We can see how our approach gets closer to the ground truth in these cases.}
    \label{fig:trajectories}
\end{figure}

We now present qualitative results from our forecasted trajectories. Fig.~\ref{fig:trajectories} shows the output of the different methods while predicting complex tracks. In these cases,  our approach yields trajectories closer to the ground truth. In the proposed self-conditioned GAN, complex trajectories are featured in small and non-dominant clusters.

\begin{figure}[t]
    \centering
        \includegraphics[width=0.48\columnwidth]{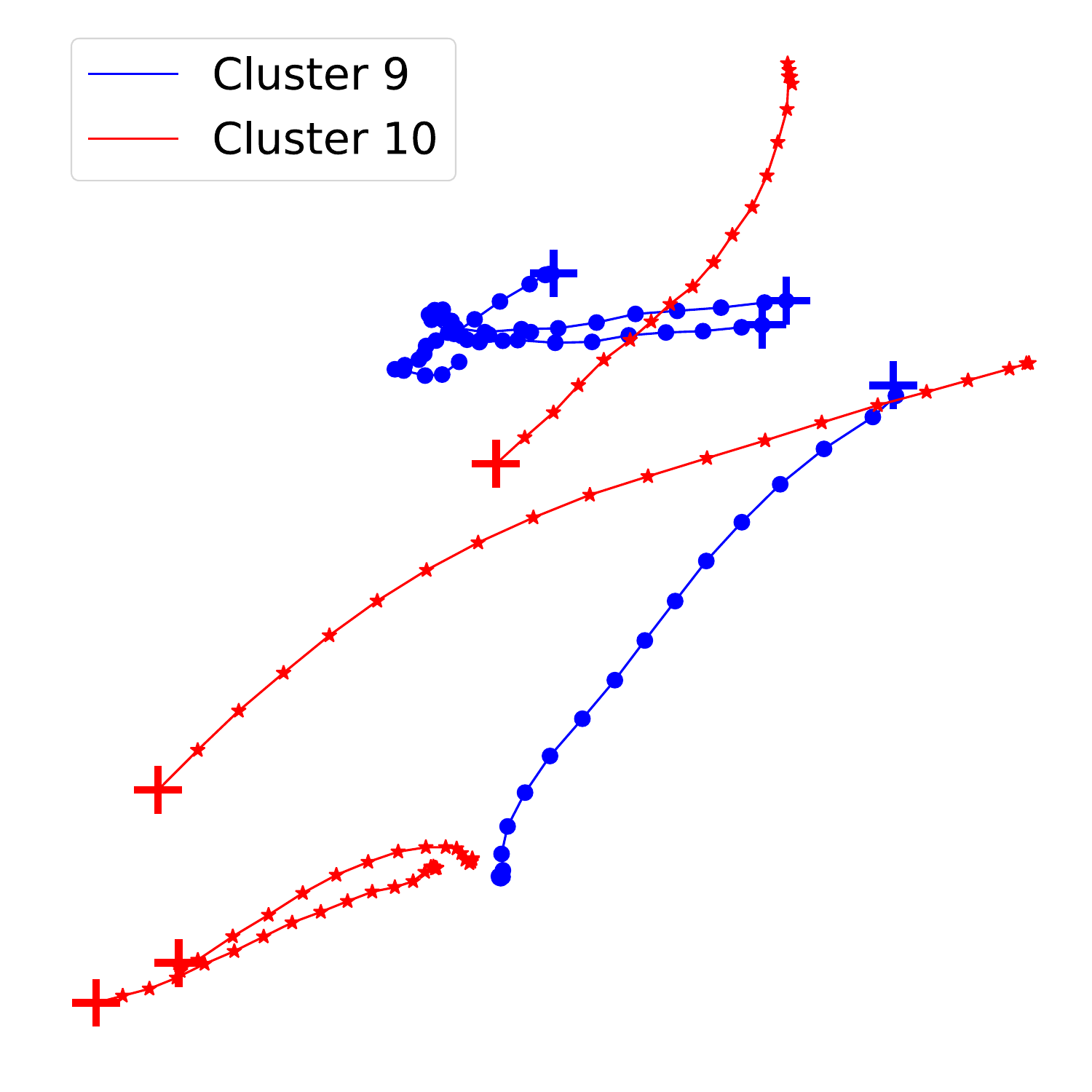}
        \includegraphics[width=0.48\columnwidth]{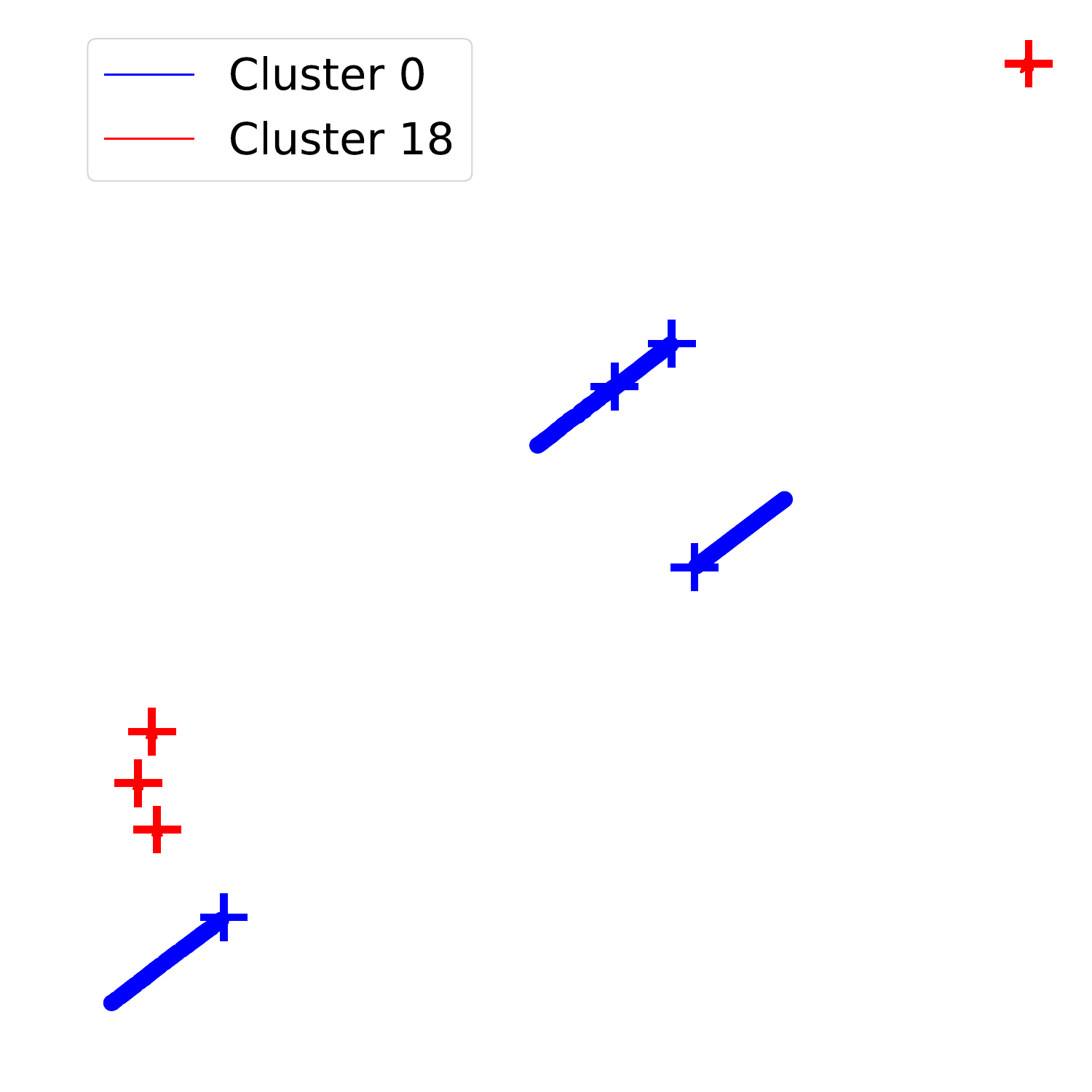}
    \caption{Entire trajectories randomly sampled from TH\"{O}R (left) and Argoverse (right) data sets, where crosses mark the starting point of each track.  In TH\"{O}R, trajectories from cluster 9 go from right to left, while tracks from cluster 10 go from left to right. In Argoverse, trajectories from cluster 0 are much longer than the ones from cluster 18.}
    \label{fig:clusters}
\end{figure}

In a second example, we present insights into the clusters resulting from the self-conditioned GAN. Our intuition is that examples belonging to the same cluster present trajectories sharing a similar behavior. In Fig.~\ref{fig:clusters}, we present the trajectories from two different clusters in both data sets. In TH\"{O}R, cluster 9 represents trajectories going from the left to the right in an ascending trend, while in cluster 10 the trajectories go in the opposite direction. Finally, in Argoverse, trajectories from cluster 0 are much longer than the ones from cluster 18.

\begin{figure}[t]
    \centering
        \includegraphics[width=0.48\columnwidth]{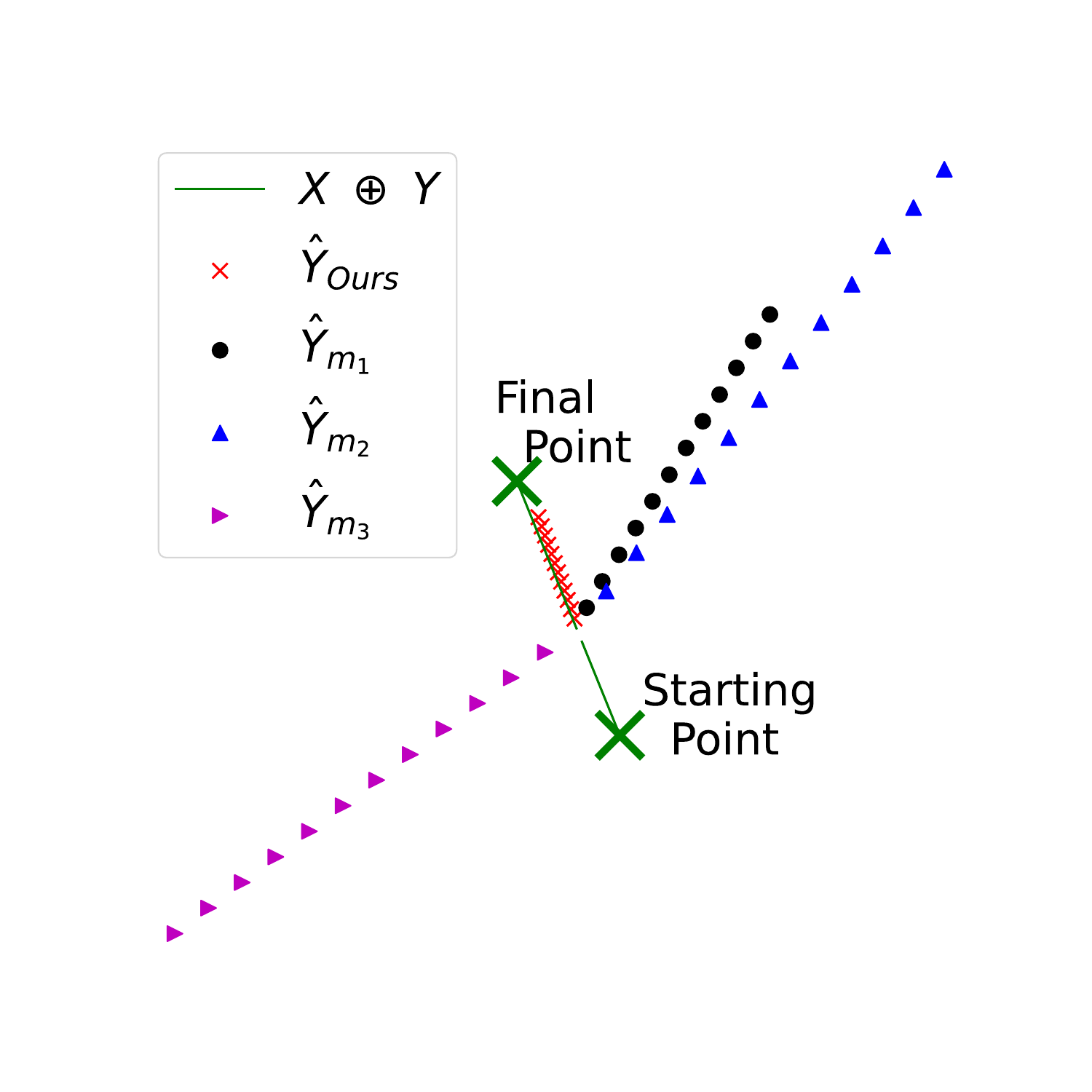}
        \includegraphics[width=0.48\columnwidth]{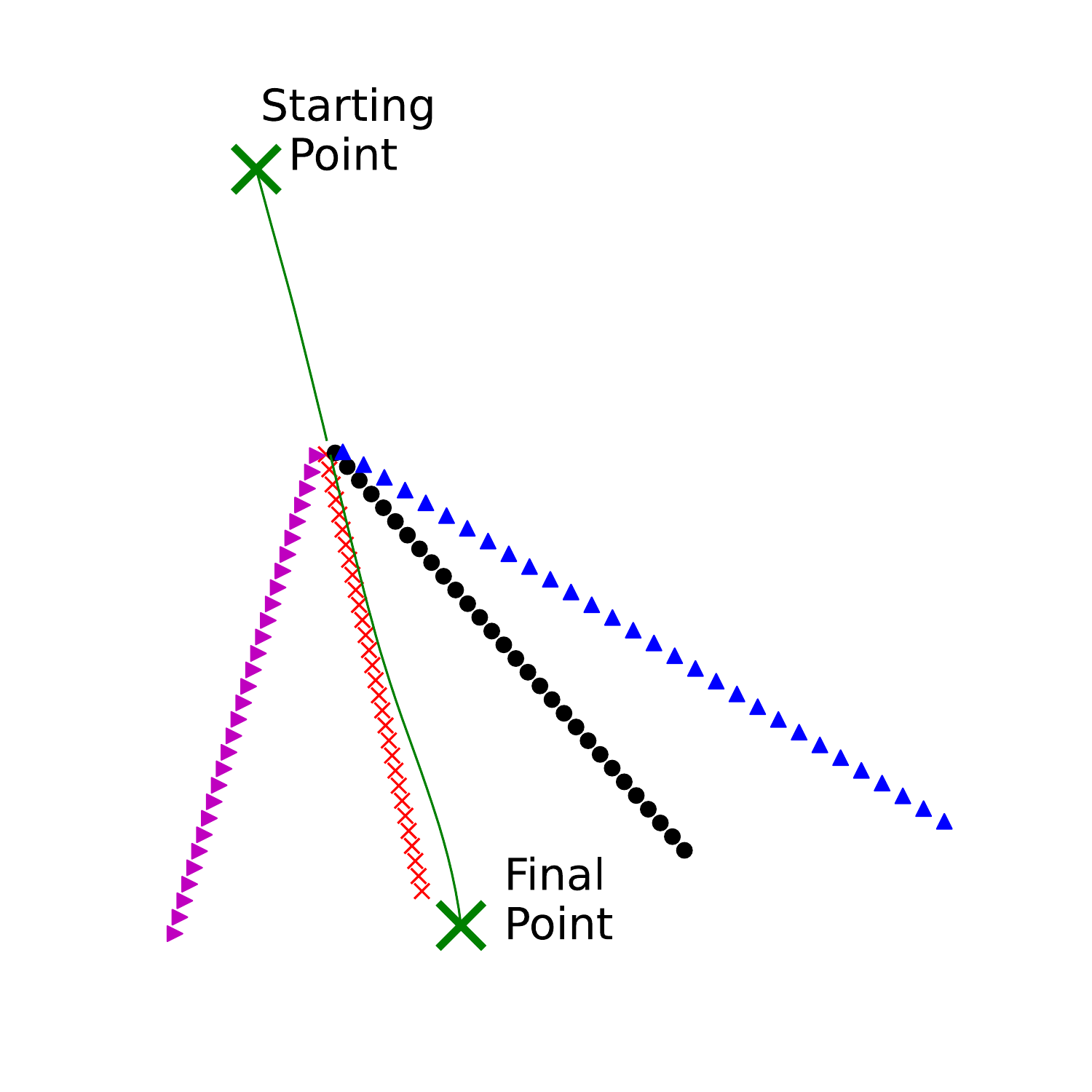}
    \caption{Test set examples of trajectories  generated from the self-conditioned GAN with four different labels in both data sets (TH\"{O}R on the left and Argoverse on the right). Both cases show that the right condition (represented by $\hat{Y}_{Ours}$) yields trajectories closer to the ground truth when compared to other three different modes ($m_{1}$, $m_{2}$, and $m_{3}$).}
    \label{fig:modes}
    \vspace{-1pt}
\end{figure}

The last example visually demonstrates the difference between the unsupervised labels in the self-conditioned method. In Fig.~\ref{fig:modes}, we present trajectories randomly sampled from clusters in the two data sets. In both cases, the correct unsupervised label is crucial to obtain more performing results. Hence, these modes are diversified and meaningful. 

\section{CONCLUSION}
In this paper we presented a context-free unsupervised approach based on a self-conditioned GAN to learn different modes from 2D trajectories and apply it to the problem of trajectory forecasting. The results showed that each mode indicates different behavioral moving patterns in the discriminator's feature space and this information improves the forecasting process. We present three training settings based on the self-conditioned GAN's clustering space, which produce better forecasters. 

We tested our method in two data sets: human motion and road agents. Experimental results showed that our approaches outperform unaware context-free methods when applied to the least representative supervised labels in these data sets, while performing well in the remaining classes. Moreover, our methods outperformed globally in human motion while performing well in road agents. Finally, we presented insights of the clustering resulting from the self-conditioned framework and the respective unsupervised labels.




\bibliographystyle{IEEEtran}
\bibliography{bib/refs}

\end{document}